\documentclass{article} 
\pdfoutput=1
\usepackage{iclr2021_conference,times}

\usepackage{hyperref}
\usepackage{url}
\usepackage{amsthm}
\usepackage{amsfonts}
\usepackage{amsmath}
\usepackage{dsfont}
\usepackage{enumitem}
	\setlist{nosep,leftmargin=*}
\usepackage{booktabs}
\usepackage{graphicx} 
	\graphicspath{{images/}}

\newcommand{\indifunc}[1]{\mathbf{1}_\mathrm{#1}}
\DeclareMathOperator{\softmax}{softmax}
\DeclareMathOperator{\id}{Id}

\newcommand{\bfx}{\mathbf{x}}
\newcommand{\bfX}{\mathbf{X}}
\newcommand{\bfT}{\mathbf{T}}
\newcommand{\bfq}{\mathbf{q}}
\newcommand{\bfe}{\mathbf{e}}
\newcommand{\bfW}{\mathbf{W}}
\newcommand{\bfU}{\mathbf{U}}
\newcommand{\bfg}{\mathbf{g}}
\newcommand{\bfv}{\mathbf{v}}
\newcommand{\bfD}{\mathbf{D}}
\newcommand{\bbE}{\mathbb{E}}
\newcommand{\pa}{Pa}

\newtheorem{atthm}{Theorem}
\newtheorem{atprop}[atthm]{Proposition}
\newtheorem{atcor}[atthm]{Corollary}
\theoremstyle{definition}
\newtheorem{atdef}{Definition}
\theoremstyle{remark}

\title{A Chain Graph Interpretation\\ of Real-World Neural Networks}

\author{Yuesong Shen, Daniel Cremers\\
Technical University of Munich\\
\texttt{\{yuesong.shen, cremers\}@tum.de}
}

\iclrfinalcopy 
\begin{document}

\maketitle

\begin{abstract}
The last decade has witnessed a boom of deep learning research and applications achieving state-of-the-art results in various domains. However, most advances have been established empirically, and their theoretical analysis remains lacking. One major issue is that our current interpretation of neural networks (NNs) as function approximators is too generic to support in-depth analysis. In this paper, we remedy this by proposing an alternative interpretation that identifies NNs as chain graphs (CGs) and feed-forward as an approximate inference procedure. The CG interpretation specifies the nature of each NN component within the rich theoretical framework of probabilistic graphical models, while at the same time remains general enough to cover real-world NNs with arbitrary depth, multi-branching and varied activations, as well as common structures including convolution / recurrent layers, residual block and dropout. We demonstrate with concrete examples that the CG interpretation can provide novel theoretical support and insights for various NN techniques, as well as derive new deep learning approaches such as the concept of partially collapsed feed-forward inference. It is thus a promising framework that deepens our understanding of neural networks and provides a coherent theoretical formulation for future deep learning research.
\end{abstract}

\section{Introduction}

During the last decade, deep learning \citep{goodfellow2016dlbook}, the study of neural networks (NNs), has achieved ground-breaking results in diverse areas such as computer vision \citep{krizhevsky12alexnet,he2016resnet,long15fully,chen2018deeplab}, natural language processing \citep{hinton2012speech,vaswani2017attention,devlin2019bert}, generative modeling \citep{kingma2014vae,goodfellow2014gan} and reinforcement learning \citep{mnih2015dqn,silver2016alphago}, and various network designs have been proposed. However, neural networks have been treated largely as ``black-box'' function approximators, and their designs have chiefly been found via trial-and-error, with little or no theoretical justification. A major cause that hinders the theoretical analysis is the current overly generic modeling of neural networks as function approximators: simply interpreting a neural network as a composition of parametrized functions provides little insight to decipher the nature of its components or its behavior during the learning process.

In this paper, we show that a neural network can actually be interpreted as a probabilistic graphical model (PGM) called chain graph (CG) \citep{koller2009pgmbook}, and feed-forward as an efficient approximate probabilistic inference on it. This offers specific interpretations for various neural network components, allowing for in-depth theoretical analysis and derivation of new approaches.

\begin{figure}[t]
\centering
\includegraphics[width=0.39\linewidth]{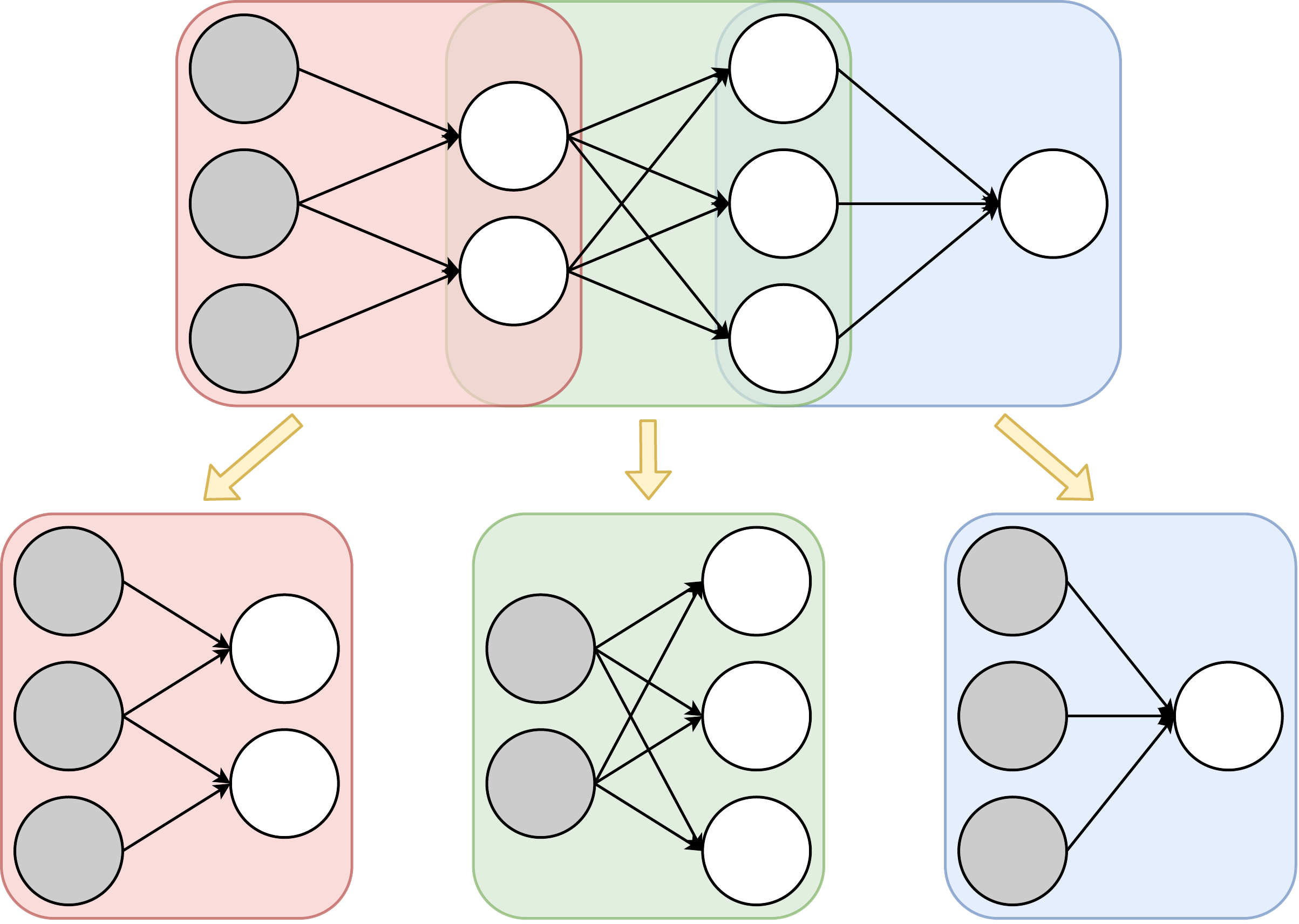}
\hspace{0.02\linewidth}
\includegraphics[width=0.39\linewidth]{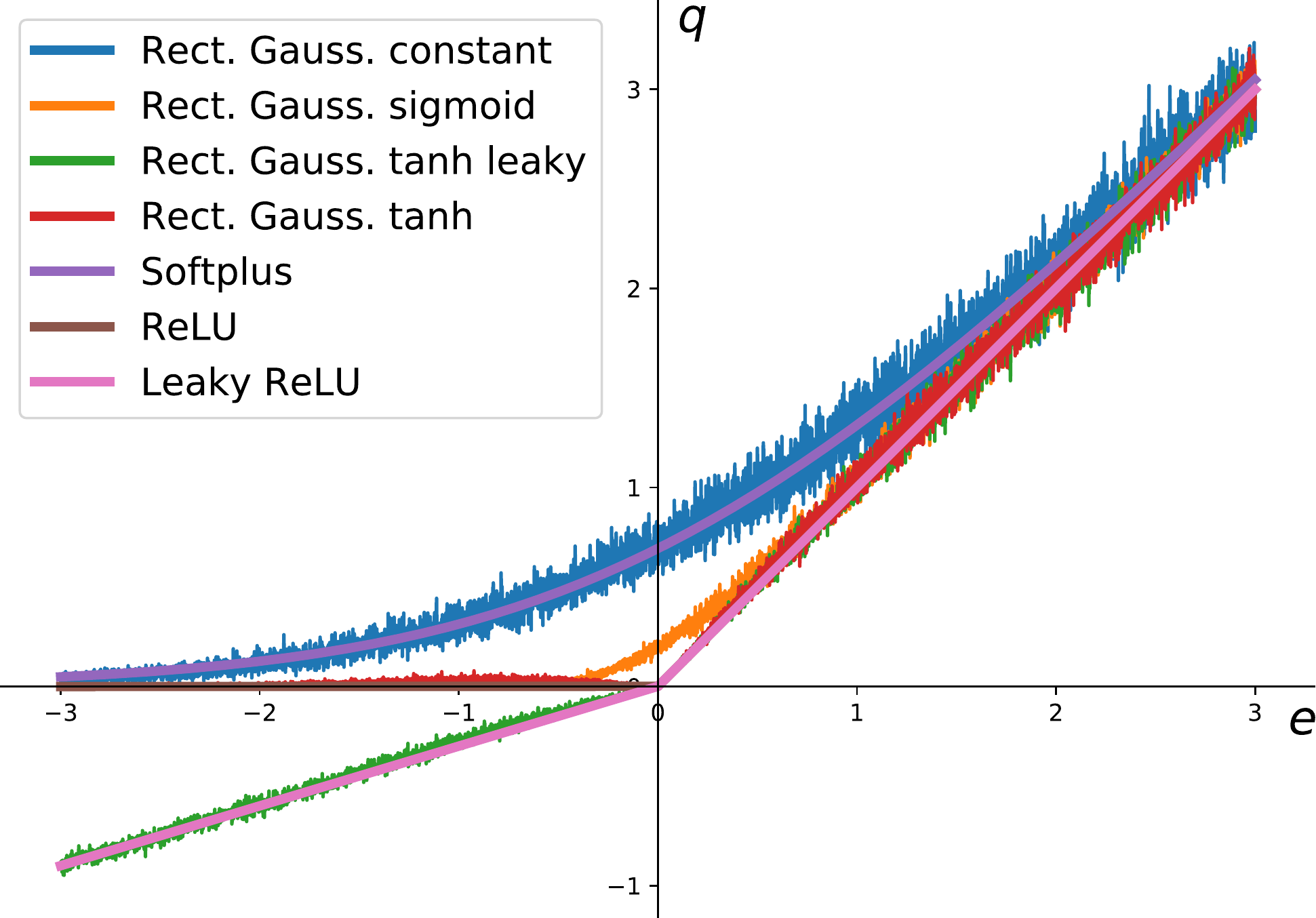}
\caption{Neural networks can be interpreted as layered chain graphs where activation functions are determined by node distributions. \textsl{Left}: An example neural network interpreted as a chain graph with three chain components which represent its layers; \textsl{Right}: A variety of activation functions (softplus, ReLU, leaky ReLU) approximated by nodes following rectified Gaussian distributions ($e,q$ as in Eq.~\eqref{eq:ff}). We visualize the approximations stochastically by averaging over 200 samples.} \label{fig:banner}
\end{figure}

\subsection{Related work} \label{subsec:relatedwork}

In terms of theoretical understanding of neural networks, a well known result based on the function approximator view is the universal approximation theorem \citep{goodfellow2016dlbook}, however it only establishes the representational power of NNs. Also, there have been many efforts on alternative NN interpretations. One prominent approach identifies infinite width NNs as Gaussian processes \citep{neal1996phd,lee2018dnnasgp}, enabling kernel method analysis \citep{jacot2018ntk}. Other works also employ theories such as optimal transport \citep{genevay2017gan,chizat2018global} or mean field \citep{mei2019mean}. These approaches lead to interesting findings, however they tend to only hold under limited or unrealistic settings and have difficulties interpreting practical real-world NNs.

Alternatively, some existing works study the post-hoc interpretability \citep{lipton2018mythos}, proposing methods to analyze the empirical behavior of trained neural networks: activation maximization \citep{erhan2009visualizing}, typical input synthesis \citep{nguyen2016synthinput}, deconvolution \citep{zeiler2014deconv},  layer-wise relevance propagation \citep{bach2015lrp}, etc. These methods can offer valuable insights to the practical behavior of  neural networks, however they represent distinct approaches and focuses, and are all limited within the function approximator view.

Our work links neural networks to probabilistic graphical models \citep{koller2009pgmbook}, a rich theoretical framework that models and visualizes probabilistic systems composed of random variables (RVs) and their interdependencies. There are several types of graphical models. The chain graph  model \citep{koller2009pgmbook} used in our work is a general form that unites directed and undirected variants, visualized as a partially directed acyclic graph (PDAG). Interestingly, there exists a series of works on constructing hierarchical graphical models for data-driven learning problems, such as sigmoid belief network \citep{neal1992sbn}, deep belief network \citep{hinton2006dbn}, deep Boltzmann machine \citep{salakhutdinov2012dbm} and sum product network \citep{poon2011spn}. As alternatives to neural networks, these models have shown promising potentials for generative modeling and unsupervised learning. Nevertheless, they are yet to demonstrate competitive performances over neural network for discriminative learning.

Neural networks and graphical models have so far been treated as two distinct approaches in general. Existing works that combine them \citep{zheng2015crfasrnn, chen2018deeplab, lampe2016crfnlp} treat either neural networks as function approximators for amortized inference, or graphical models as post-processing steps. Some consider neural networks as graphical models with deterministic hidden nodes \citep{buntine1994gmasnn}. However this is an atypical degenerate regime. To the best of our knowledge, our work provides the first rigorous and comprehensive formulation of a (non-degenerate) graphical model interpretation for neural networks in practical use.

\subsection{Our contributions}

The main contributions of our work are summarized as follows:
\begin{itemize}
\item We propose a layered chain graph representation of neural networks, interpret feed-forward as an approximate probabilistic inference procedure, and show that this interpretation provides an extensive coverage of practical NN components (Section~\ref{sec:nnascg});
\item To illustrate its advantages, we show with concrete examples (residual block, RNN, dropout) that the chain graph interpretation enables coherent and in-depth theoretical support,  and provides additional insights to various empirically established network structures (Section~\ref{sec:case_studies});
\item Furthermore, we demonstrate the potential of the chain graph interpretation for discovering new approaches by using it to derive a novel stochastic inference method named partially collapsed feed-forward, and establish experimentally its empirical effectiveness (Section~\ref{sec:pcff}).
\end{itemize}

\section{Chain graph interpretation of neural networks}
\label{sec:nnascg}

Without further delay, we derive the chain graph interpretation of neural networks in this section. We will state and discuss the main results here and leave the proofs in the appendix.

\subsection{The layered chain graph representation}
\label{subsec:cgrepr}

We start by formulating the so called \textsl{layered chain graph} that corresponds to neural networks we use in practice: Consider a system represented by $L$ layers of random variables $(\bfX^{1}, \dots, \bfX^{L})$, where $X^{l}_i$ is the $i$-th variable node in the $l$-th layer, and denote $N^l$ the number of nodes in layer $l$. We assume that nodes $X^{l}_i$ in the same layer $l$ have the same distribution type characterized by a feature function $\bfT^l$ that can be multidimensional. Also, we assume that the layers are ordered topologically and denote $\pa(\bfX^l)$ the parent layers of $\bfX^l$. To ease our discussion, we assume that $\bfX^1$ is the input layer and $\bfX^L$ the output layer (our formulation can easily extend to multi-input/output cases). A layered chain graph is then defined as follows:

\begin{atdef}
A \textsl{layered chain graph} that involves $L$ layers of random variables $(\bfX^{1}, \dots, \bfX^{L})$ is a chain graph that encodes the overall distribution $P(\bfX^2, \dots, \bfX^L | \bfX^1)$ such that:
\begin{enumerate}
\item It can be factored into layerwise chain components $P(\bfX^{l}|\pa(\bfX^l))$ following the topological order, and nodes $X^{l}_i$ within each chain component $P(\bfX^{l}|\pa(\bfX^l))$  are conditionally independent given their parents (this results in bipartite chain components), thus allowing for further decomposition into nodewise conditional distributions $P(X^{l}_i|\pa(\bfX^l))$ . This means we have
\begin{equation}
P(\bfX^2, \dots, \bfX^L | \bfX^1) = \prod_{l=2}^{L} P(\bfX^{l}|\pa(\bfX^l)) = \prod_{l=2}^{L} \prod_{i=1}^{N^{l}} P(X_i^{l}|\pa(\bfX^l));
\end{equation}

\item For each layer $l$ with parent layers $\pa(\bfX^l)=\{ \bfX^{p_1}, \dots \bfX^{p_n} \}, p_1,\dots,p_n \in \{ 1,\dots,l-1 \}$, its nodewise conditional distributions $P(X^{l}_i|\pa(\bfX^l))$ are modeled by pairwise conditional random fields (CRFs) with with unary ($\mathbf{b}^{l}_i$) and pairwise ($\bfW_{j, i}^{p, l}$) weights (as we will see, they actually correspond to biases and weights in NN layers):
\begin{align}
&P(X^{l}_i|\pa(\bfX^l)) = f^{l} \big( \bfT^l(X^{l}_i), \bfe^{l}_i\big(\bfT^{p_1}(\bfX^{p_1}),\dots,\bfT^{p_n}(\bfX^{p_n})\big) \big) \label{eq:activgeneral} \\
\text{with}\quad &\bfe^{l}_i\big(\bfT^{p_1}(\bfX^{p_1}),\dots,\bfT^{p_n}(\bfX^{p_n})\big) = \mathbf{b}^{l}_i + \sum_{p=p_1}^{p_n} \sum_{j=1}^{N^{p}} \bfW_{j, i}^{p, l} \bfT^{p}(X^{p}_j). \label{eq:activgeneralpreact}
\end{align}
\end{enumerate}
\end{atdef}

Figure~\ref{fig:banner}~Left illustrates an example three-layer network as layered chain graph and its chain component factorization. For exponential family distributions \citep{koller2009pgmbook}, the general form in Eq.~\eqref{eq:activgeneral} simply becomes $P(X^{l}_i|\pa(\bfX^l)) \propto \exp \big( \bfT^l(X^{l}_i) \cdot \bfe^{l}_i\big(\bfT^{p_1}(\bfX^{p_1}),\dots,\bfT^{p_n}(\bfX^{p_n})\big) \big)$.

\subsection{Feed-forward as approximate probabilistic inference} \label{subsec:ffprobainfer}

To identify layered chain graphs with real-world neural networks, we need to show that they can behave the same way during inference and learning. For this, we establish the fact that feed-forward can actually be seen as performing an approximate probabilistic inference on a layered chain graph: Given an input sample $\tilde{\bfx}^1$, we consider the problem of inferring the marginal distribution $Q^l_i$ of a node $X^l_i$ and its expected features $\bfq^l_i$, defined as
\begin{equation}
Q^l_i (x^l_i | \tilde{\bfx}^1) = P(X^l_i=x^l_i | \bfX^1=\tilde{\bfx}^1);\quad \bfq^l_i = \bbE_{Q^l_i}[\bfT^l(X^l_i)] \enspace (
\bfq^1 = \tilde{\bfx}^1). \label{eq:qdefs}
\end{equation}
Consider a non-input layer $l$ with parent layers $p_1,\dots,p_n$, the independence assumptions encoded by the layered chain graph lead to the following recursive expression for marginal distributions $Q$:
\begin{equation}
Q^l_i (x^l_i | \tilde{\bfx}^1) =  \bbE_{Q^{p_1},\dots,Q^{p_n}}[P(x^l_i | \pa(\bfX^l))]. \label{eq:qrecursive}
\end{equation}
However, the above expression is in general intractable, as it integrates over the entire admissible states of all parents nodes in $\pa(\bfX^l)$. To proceed further, simplifying approximations are needed. Interestingly, by using linear approximations, we can obtain the following results (in case of discrete random variable the integration in Eq.~\ref{eq:ff} is replaced by summation):

\begin{atprop} \label{thm:ff}
If we make the assumptions that the corresponding expressions are approximately linear w.r.t.\ parent features $\bfT^{p_1}(\bfX^{p_1}),\dots,\bfT^{p_n}(\bfX^{p_n})$, we obtain the following approximations:
\begin{align}
&Q^l_i (x^l_i | \tilde{\bfx}^1) \approx f^{l} \big( \bfT^l(x^{l}_i), \bfe^{l}_i(\bfq^{p_1},\dots,\bfq^{p_n}) \big); \label{eq:distff} \\
&\bfq^l_i \approx \int_{x^l_i} \bfT^l(x^l_i) f^{l} \big( \bfT^l(x^{l}_i), \bfe^{l}_i(\bfq^{p_1},\dots,\bfq^{p_n}) \big) dx^l_i := \bfg^l(\bfe^{l}_i(\bfq^{p_1},\dots,\bfq^{p_n})). \label{eq:ff}
\end{align}
Especially, Eq.~\eqref{eq:ff} is a feed-forward expression for expected features $\bfq^l_i$ with activation function $\bfg^l$ determined by $\bfT^l$ and $f^l$, i.e.\ the distribution type of random variable nodes in layer $l$. 
\end{atprop}

The proof is provided in Appendix~\ref{asubsec:proof_thm_ff}. This allows us to identify feed-forward as an approximate probabilistic inference procedure for layered chain graphs. For learning, the loss function is typically a function of $(Q^L,\bfq^L)$ obtainable via feed-forward, and we can follow the same classical neural network parameter update using stochastic gradient descent and backpropagation. Thus we are able to replicate the exact neural network training process with this layered chain graph framework.

The following corollary provides concrete examples of some common activation functions $\bfg$ (we emphasize their names in bold, detailed formulations and proofs are given in Appendix~\ref{asubsec:proof_cor_activationfuncs}):

\begin{atcor} \label{cor:activationfuncs}
We have the following node distribution \-- activation function correspondences:
\begin{enumerate}
\item Binary nodes taking values $\{\alpha, \beta\}$ results in sigmoidal activations, especially, we obtain \textbf{sigmoid} with $\alpha=0, \beta=1$ and \textbf{tanh} with $\alpha=-1, \beta=1$ ($\alpha, \beta$ are interchangeable);

\item Multilabel nodes characterized by label indicator features result in the \textbf{softmax} activation; 

\item Variants of (leaky) rectified Gaussian distributions ($T^{l}_i(X^{l}_i) = X^{l}_i = \max(\epsilon Y^{l}_i, Y^{l}_i)$ with $Y^{l}_i \sim \mathcal{N} \big( e^{l}_i, (s^l_i(e^{l}_i))^2 \big)$) can approximate activations such as \textbf{softplus} ($\epsilon=0, s^l_i \approx 1.7761$) and \textbf{$\epsilon$-leaky rectified linear unit} (ReLU) ($s^l_i =\tanh (e^{l}_i)$) including \textbf{ReLU} ($\epsilon=0$) and \textbf{identity} ($\epsilon=1$).
\end{enumerate}
\end{atcor}

Figure~\ref{fig:banner}~Right illustrates activation functions approximated by various rectified Gaussian variants. We also plotted (in orange) an alternative approximation of ReLU with sigmoid-modulated standard deviation proposed by \citet{nair2010relu} which is less accurate around the kink at the origin.

The linear approximations, needed for feed-forward, is coarse and only accurate for small pairwise weights ($\|\bfW\| \ll 1$) or already linear regions. This might justify weight decay beyond the general ``anti-overfit'' argument and the empirical superiority of piecewise linear activations like ReLU \citep{nair2010relu}.  Conversely, as a source of error, it might explain some ``failure cases'' of neural networks such as their vulnerability against adversarial samples, see e.g.,\ \citet{goodfellow2015adversarial}.

\subsection{Generality of the chain graph interpretation} \label{subsec:gen_nnascg}

The chain graph interpretation formulated in Sections~\ref{subsec:cgrepr} and \ref{subsec:ffprobainfer} is a general framework that can describe many practical network structures. To demonstrate this, we list here a wide range of neural network designs (marked in bold) that are chain graph interpretable.

\begin{itemize}
\item In terms of network architecture, it is clear that the chain graph interpretation can model networks of arbitrary depth, and with general multi-branched structures  such as \textbf{inception modules} \citep{szegedy2015googlenet} or \textbf{residual blocks} \citep{he2016resnet, he2016resnet2} discussed in Section~\ref{subsec:residual}. Also, it is possible to built up \textbf{recurrent neural networks (RNNs)} for sequential data learning, as we will see in Section~\ref{subsec:rnn}. Furthermore,  the modularity of chain components justifies \textbf{transfer learning via partial reuse of pre-trained networks}, e.g.,\ backbones trained for image classification can be reused for segmentation \citep{chen2018deeplab}.

\item In terms of layer structure, we are free to employ sparse connection patterns and shared/fixed weight, so that we can obtain not only \textbf{dense connections}, but also connections like \textbf{convolution}, \textbf{average pooling} or \textbf{skip connections}. Moreover, as shown in Section~\ref{subsec:dropout}, \textbf{dropout} can be reproduced by introducing and sampling from auxiliary random variables, and normalization layers like \textbf{batch normalization} \citep{ioffe2015batchnorm} can be seen as reparametrizations of node distributions and fall within the general form (Eq.~\eqref{eq:activgeneral}). Finally, we can extend the layered chain graph model to allow for intra-layer connections, which enables \textbf{non-bipartite CRF layers} which are typically used on output layers for structured prediction tasks like image segmentation \citep{zheng2015crfasrnn, chen2018deeplab} or named entity recognition \citep{lampe2016crfnlp}. However, feed-forward is no longer applicable through these intra-connected layers.

\item Node distributions can be chosen freely, leading to a variety of nonlinearities (e.g.,\ Corollary~\ref{cor:activationfuncs}).
\end{itemize}

\section{Selected case studies of existing neural network designs} \label{sec:case_studies}

The proposed chain graph interpretation offers a detailed description of the underlying mechanism of neural networks. This allows us to obtain novel theoretical support and insights for various network designs which are consistent within a unified framework. We illustrate this with the following concrete examples where we perform in-depth analysis based on the chain graph formulation.

\subsection{Residual block as refinement module} \label{subsec:residual}

The residual block, proposed originally in \citet{he2016resnet} and improved later \citep{he2016resnet2} with the preactivation form, is an effective design for building up very deep networks. Here we show that a preactivation residual block corresponds to a refinement module within a chain graph. We use \textsl{modules} to refer to encapsulations of layered chain subgraphs as input--output mappings without specifying their internal structures. A refinement module is defined as follows:

\begin{atdef} \label{def:refinementmodule}
Given a base submodule from layer $\bfX^{l-1}$ to layer $\bfX^l$, a \textsl{refinement module} augments this base submodule with a side branch that chains a copy of the base submodule (sharing weight with its original) from $\bfX^{l-1}$ to a duplicated layer $\tilde{\bfX}^l$, and then a refining submodule from $\tilde{\bfX}^l$ to $\bfX^l$.
\end{atdef}

\begin{figure}[ht]
\centering
\includegraphics[width=0.24\linewidth]{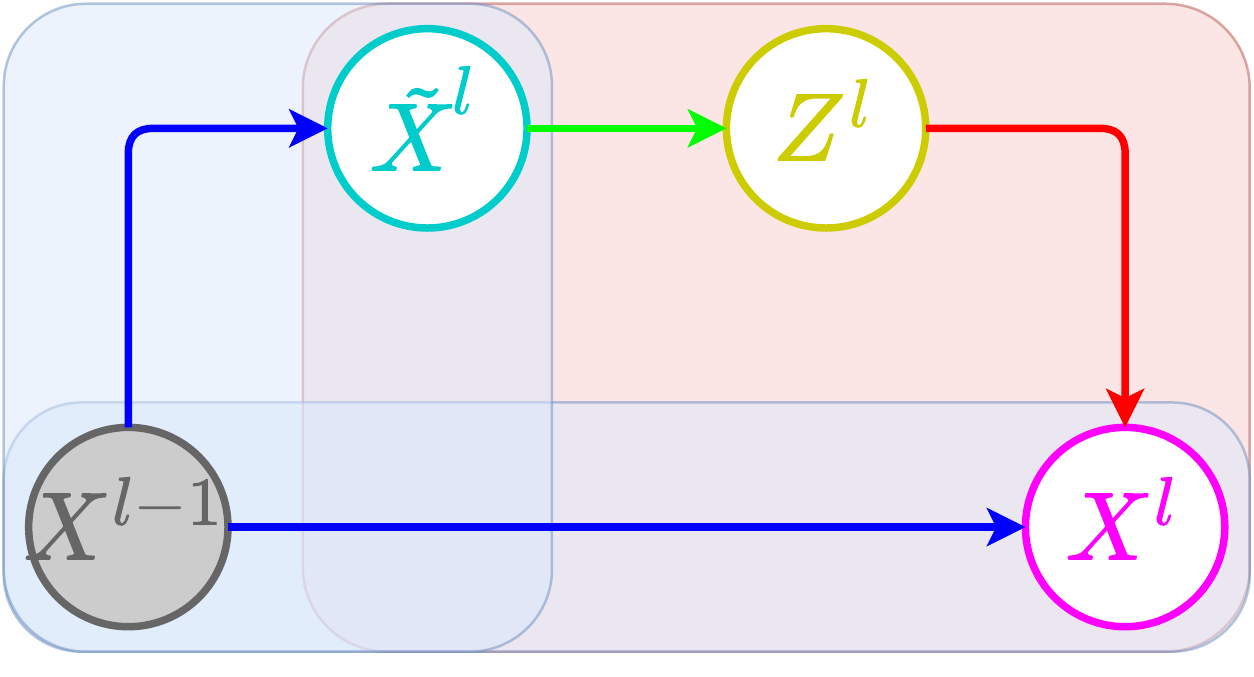}
\hspace{0.02\linewidth}
\includegraphics[width=0.6\linewidth]{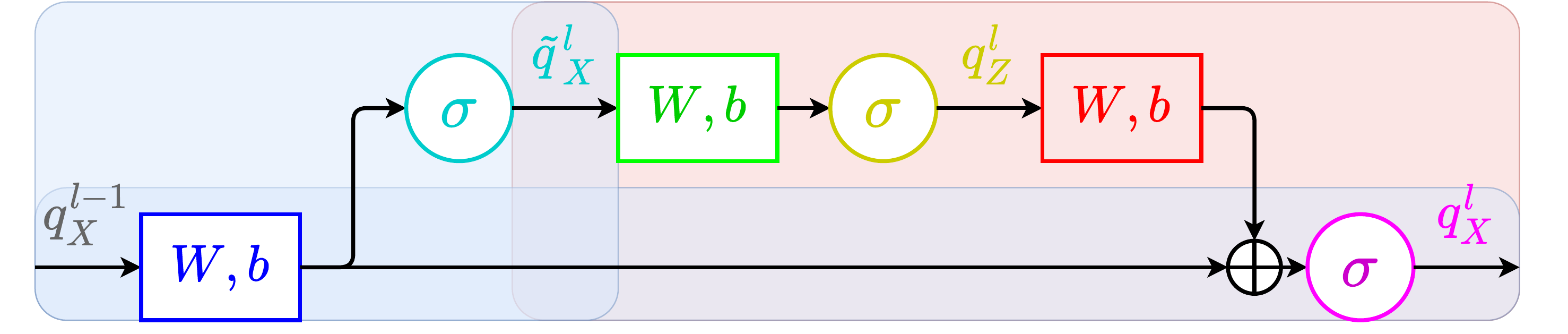}
\caption{Example of a refinement module (left) and its corresponding computational graph (right), composed of a base submodule $X^{l-1} \to X^l$ (blue background) and a refining submodule $\tilde{X}^l \to Z^l \to X^l$ (red background). In the computational graph each $W,b$ represents a linear connection (Eq.~\eqref{eq:activgeneralpreact}) and $\sigma$ an activation function. Same color identifies corresponding parts in the two graphs. We see that this refinement module corresponds exactly to a preactivation residual block.} \label{fig:residualblock}
\end{figure}

\begin{atprop} \label{prop:residual}
A refinement module corresponds to a preactivation residual block.
\end{atprop}

We provide a proof in Appendix~\ref{asubsec:proof_prop_residual} and illustrate this correspondence in Figure~\ref{fig:residualblock}. An interesting remark is that the refinement process can be recursive: the base submodule of a refinement module can be a refinement module itself. This results in a sequence of consecutive residual blocks.

While a vanilla layered chain component encodes a generalized linear model during feed-forward (c.f.\ Eqs.~\eqref{eq:ff},\eqref{eq:activgeneralpreact}), the refinement process introduces a nonlinear extension term to the previously linear output preactivation, effectively increasing the representational power. This provides a possible explanation to the empirical improvement generally observed when using residual blocks.

Note that it is also possible to interpret the original postactivation residual blocks, however in a somewhat artificial manner, as it requires defining identity connections with manually fixed weights.

\subsection{Recurrent neural networks} \label{subsec:rnn}

Recurrent neural networks (RNNs) \citep{goodfellow2016dlbook} are widely used for handling sequential data. An unrolled recurrent neural network can be interpreted as a dynamic layered chain graph constructed as follows: a given base layered chain graph is copied for each time step, then these copies are connected together through recurrent chain components following the Markov assumption \citep{koller2009pgmbook}: each recurrent layer $\bfX^{l,t}$ at time $t$ is connected by its corresponding layer $\bfX^{l,t-1}$ from the previous time step $t-1$. Especially, denoting $\pa^t(\bfX^{l,t})$ the non-recurrent parent layers of $\bfX^{l,t}$ in the base chain graph, we can easily interpret the following two variants: 

\begin{atprop} \label{prop:rnn} Given a recurrent chain component that encodes $P(\bfX^{l,t}|\pa^t(\bfX^{l,t}), \bfX^{l,t-1})$,
\begin{enumerate}
\item It corresponds to a simple (or vanilla / Elman) recurrent layer \citep{goodfellow2016dlbook} if the connection from $\bfX^{l,t-1}$ to $\bfX^{l,t}$ is dense;
\item It corresponds to an independently RNN (IndRNN) \citep{li2018indrnn} layer if the conditional independence assumptions among the nodes $X^{l,t}_i$ within layer $l$ are kept through time:
\begin{equation}
\forall i \in \{1,\dots, N^l\}, \ P(X^{l,t}_i|\pa^t(\bfX^{l,t}), \bfX^{l,t-1}) = P(X^{l,t}_i|\pa^t(\bfX^{l,t}), X^{l,t-1}_i).
\end{equation}
\end{enumerate}
\end{atprop}

\begin{figure}[ht]
\centering
\includegraphics[width=0.35\linewidth]{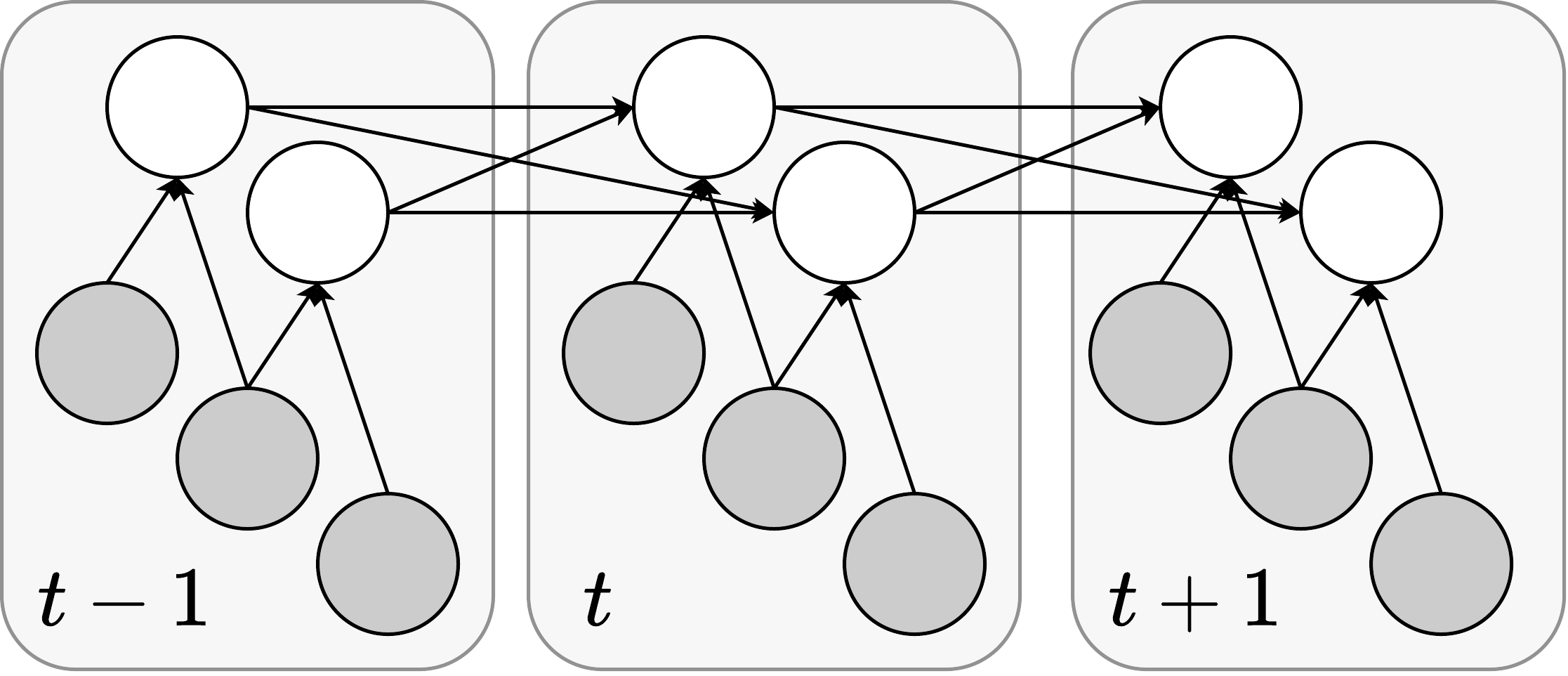}
\hspace{0.06\linewidth}
\includegraphics[width=0.35\linewidth]{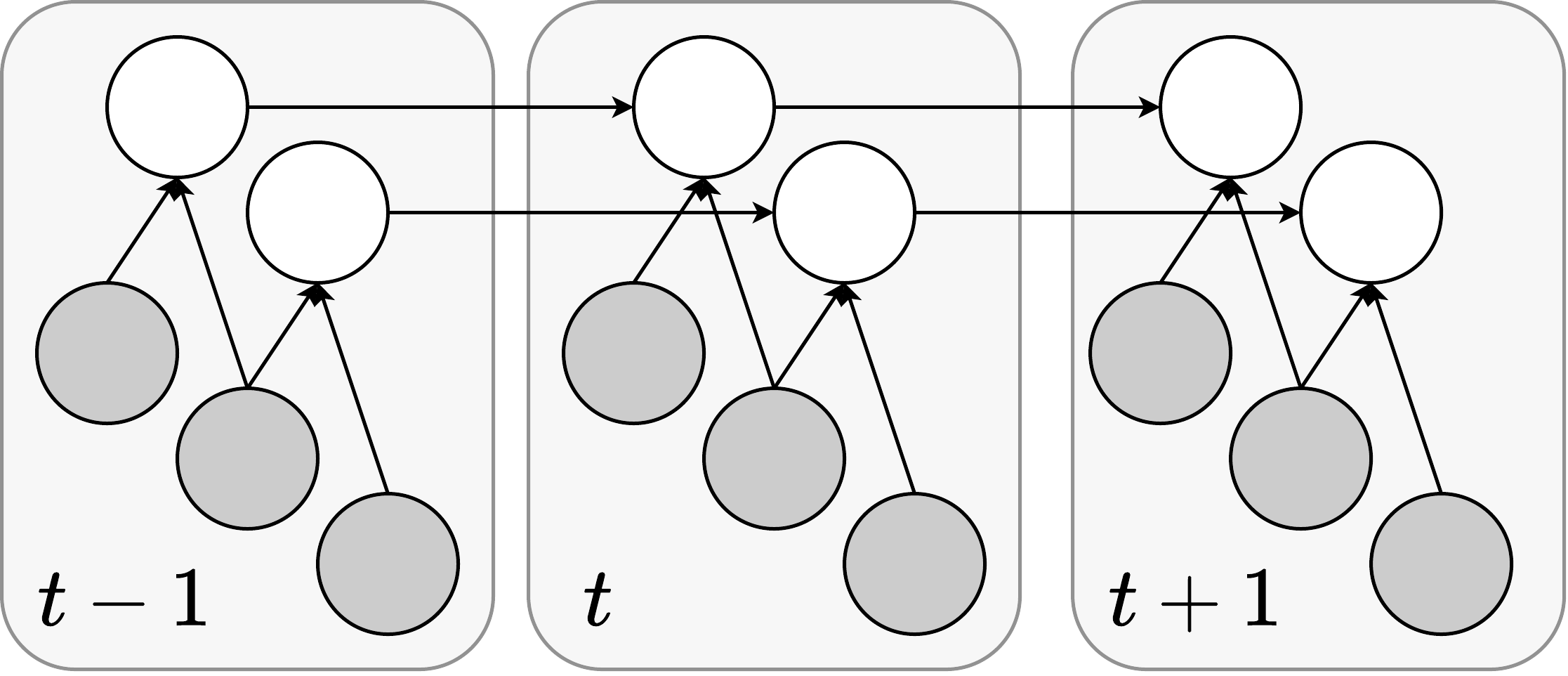}
\caption{Comparison of an simple recurrent layer (left) v.s.\ an IndRNN (right) recurrent layer. IndRNN, the better variant, enforces the intra-layer conditional independence through time.} \label{fig:rnn}
\end{figure}

We provide a proof in Appendix~\ref{asubsec:proof_prop_rnn} and illustrates both variants in Figure~\ref{fig:rnn}.

The simple recurrent layer, despite its exhaustive dense recurrent connection, is known to suffer from vanishing/exploding gradient and can not handle long sequences. The commonly used long-short term memory \citep{hochreiter1997lstm} and gated recurrent unit \citep{cho2014gru} alleviate this issue via long term memory cells and gating. However, they tend to result in bloated structures, and still cannot handle very long sequences \citep{li2018indrnn}. On the other hand, IndRNNs can process much longer sequences and significantly outperform not only simple RNNs, but also LSTM-based variants \citep{li2018indrnn,li2019indrnn}. This indicates that the assumption of intra-layer conditional independence through time, analogue to the local receptive fields of convolutional neural networks, could be an essential sparse network design tailored for sequential modeling.

\subsection{Dropout} \label{subsec:dropout}

Dropout \citep{srivastava2014dropout} is a practical stochastic regularization method commonly used especially for regularizing fully connected layers. As we see in the following proposition, from the chain graph point of view, dropout corresponds to introducing Bernoulli auxiliary random variables that serve as noise generators for feed-forward during training:

\begin{atprop} \label{prop:dropout}
Adding dropout with drop rate $1 - p^l$ to layer $l$ corresponds to the following chain graph construction: for each node $X^l_i$ in layer $l$ we introduce an auxiliary Bernoulli random variable $D^l_i \sim \text{Bernoulli}(p^l)$ and multiply it with the pairwise interaction terms in all preactivations (Eq.~\eqref{eq:activgeneralpreact}) involving $X^l_i$ as parent (this makes $D^l_i$ a parent of all child nodes of $X^l_i$ and extend their pairwise interactions with $X^l_i$ to ternary ones). The behavior of dropout is reproduced exactly if:
\begin{itemize}
\item During training, we sample auxiliary nodes $D^l_i$ during each feed-forward. This results in dropping each activation $\bfq^l_i$ of node $X^l_i$ with probability $1 - p^l$;
\item At test time, we marginalize auxiliary nodes $D^l_i$ during each feed-forward. This leads to deterministic evaluations with a constant scaling of $p^l$ for the node activations $\bfq^l_i$.
\end{itemize}
\end{atprop}

We provide a proof in Appendix~\ref{asubsec:proof_prop_dropout}. Note that among other things, this chain graph interpretation of dropout provides a theoretical justification of the constant scaling at test time. This was originally proposed as a heuristic in \citet{srivastava2014dropout} to maintain consistent behavior after training.

\section{Partially collapsed feed-forward} \label{sec:pcff}

The theoretical formulation provided by the chain graph interpretation can also be used to derive new approaches for neural networks. It allows us to create new deep learning methods following a coherent framework that provides specific semantics to the building blocks of neural networks. Moreover, we can make use of the abundant existing work from the PGM field, which also serves as a rich source of inspiration. As a concrete example, we derive in this section a new stochastic inference procedure called partially collapsed feed-forward (PCFF) using the chain graph formulation.

\subsection{PCFF: chain graph formulation} \label{subsec:pcffcg}

A layered chain graph, which can represent a neural network, is itself a probabilistic graphical model that encodes an overall distribution conditioned on the input. This means that, to achieve stochastic behavior, we can directly draw samples from this distribution, instead of introducing additional ``noise generators'' like in dropout. In fact, given the globally directed structure of layered chain graph, and the fact that the conditioned input nodes are ancestral nodes without parent, it is a well-known PGM result that we can apply forward sampling (or ancestral sampling) \citep{koller2009pgmbook} to efficiently generate samples: given an input sample $\tilde{\bfx}^1$, we follow the topological order and sample each non-input node $X^l_i$ using its nodewise distribution (Eq.~\eqref{eq:activgeneral}) conditioned on the samples $(\bfx^{p_1}, \dots,\bfx^{p_n})$ of its parents. Compared to feed-forward, forward sampling also performs a single forward pass, but generates instead an unbiased stochastic sample estimate.

While in general an unbiased estimate is preferable and the stochastic behavior can also introduce regularization during training \citep{srivastava2014dropout}, forward sampling can not directly replace feed-forward, since the sampling operation is not differentiable and will jeopardize the gradient flow during backpropagation. To tackle this, one idea is to apply the reparametrization trick \citep{kingma2014vae} on continuous random variables (for discrete RVs the Gumbel softmax trick \citep{jang2017gumbelsoftmax} can be used but requires additional continuous relaxation). An alternative solution is to only sample part of the nodes as in the case of dropout.

The proposed partially collapse feed-forward follows the second idea: we simply ``mix up'' feed-forward and forward sampling, so that for each forward inference during training, we randomly select a portion of nodes to sample and the rest to compute deterministically with feed-forward. Thus for a node $X_i^l$ with parents $(\bfX^{p_1},\dots,\bfX^{p_n})$, its forward inference update becomes 
\begin{equation}
\bfq_i^l \gets
\begin{cases}
\bfg^l(\bfe^l_i(\bfq^{p_1}, \dots, \bfq^{p_n})) &\text{ if collapsed (feed-forward)}; \\
\bfT^l(x_i^l),\ x_i^l \sim f^{l} \big( \bfT^l(X^{l}_i), \bfe^{l}_i(\bfq^{p_1},\dots,\bfq^{p_n}) \big) &\text{ if uncollapsed (forward sampling)}.
\end{cases}
\end{equation}
Following the collapsed sampling \citep{koller2009pgmbook} terminology, we call this method the partially collapsed feed-forward (PCFF). PCFF is a generalization over feed-forward and forward sampling, which can be seen as its fully collapsed / uncollapsed extremes. Furthermore, it offers a bias--variance trade-off, and can be combined with the reparametrization trick to achieve unbiased estimates with full sampling, while simultaneously maintaining the gradient flow.

\subsection{PCFF: experimental validation} \label{subsec:pcffexp}

In the previous sections, we have been discussing existing approaches whose empirical evaluations have been thoroughly covered by prior work. The novel PCFF approach proposed in this section, however, requires experiments to check its practical effectiveness. For this we conduct here a series of experiments\footnote{Implementation available at: \url{https://github.com/tum-vision/nnascg}}. Our emphasis is to understand the behavior of PCFF under various contexts and not to achieve best result for any specific task. We only use chain graph interpretable components, and we adopt the reparameterization trick  \citep{kingma2014vae} for ReLU PCFF samples.

The following experiments show that PCFF is overall an effective stochastic regularization method. Compared to dropout, it tends to produce more consistent performance improvement, and can sometimes outperform dropout. This confirms that our chain graph based reasoning has successfully found an interesting novel deep learning method.

\begin{figure}[ht]
\centering
\includegraphics[width=0.47\linewidth]{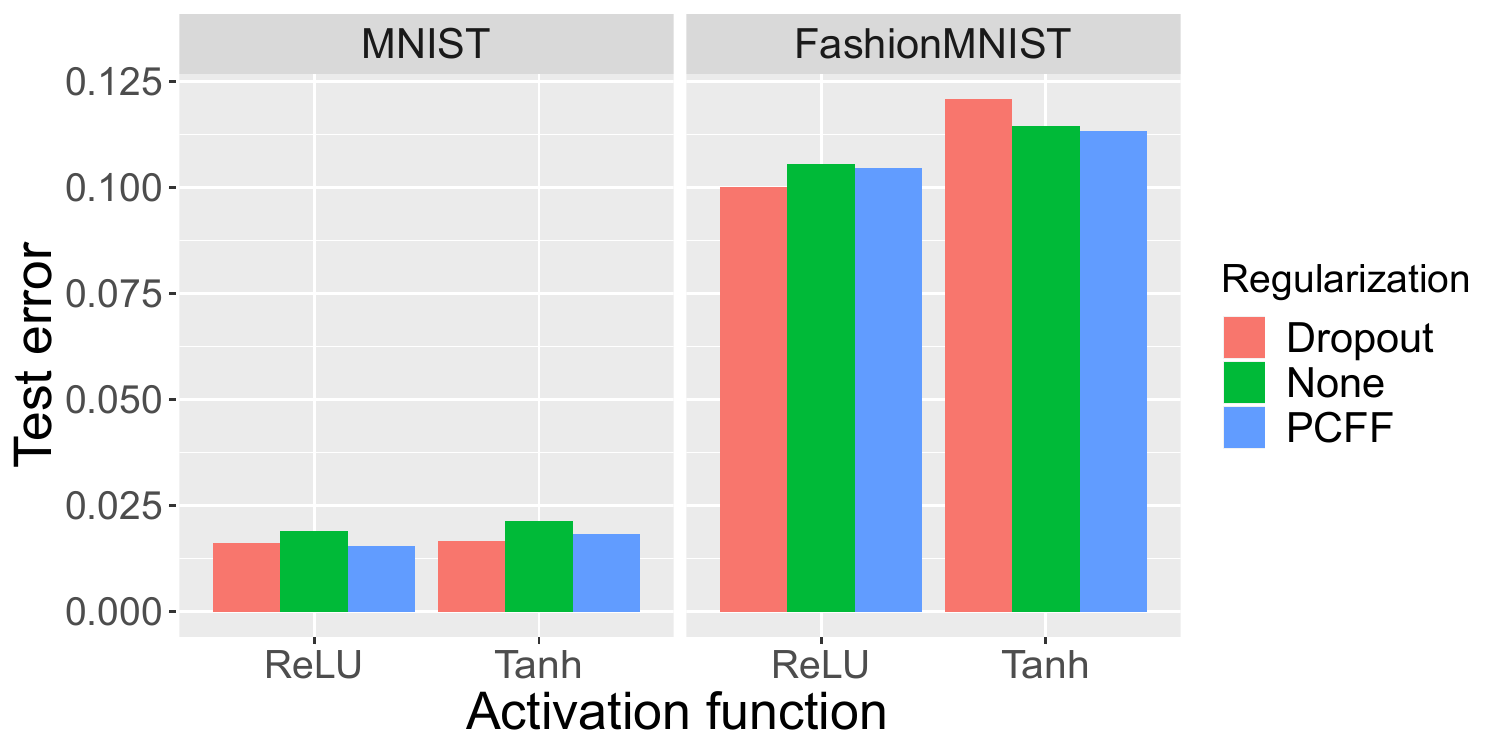}
\hspace{0.04\linewidth}
\includegraphics[width=0.36\linewidth]{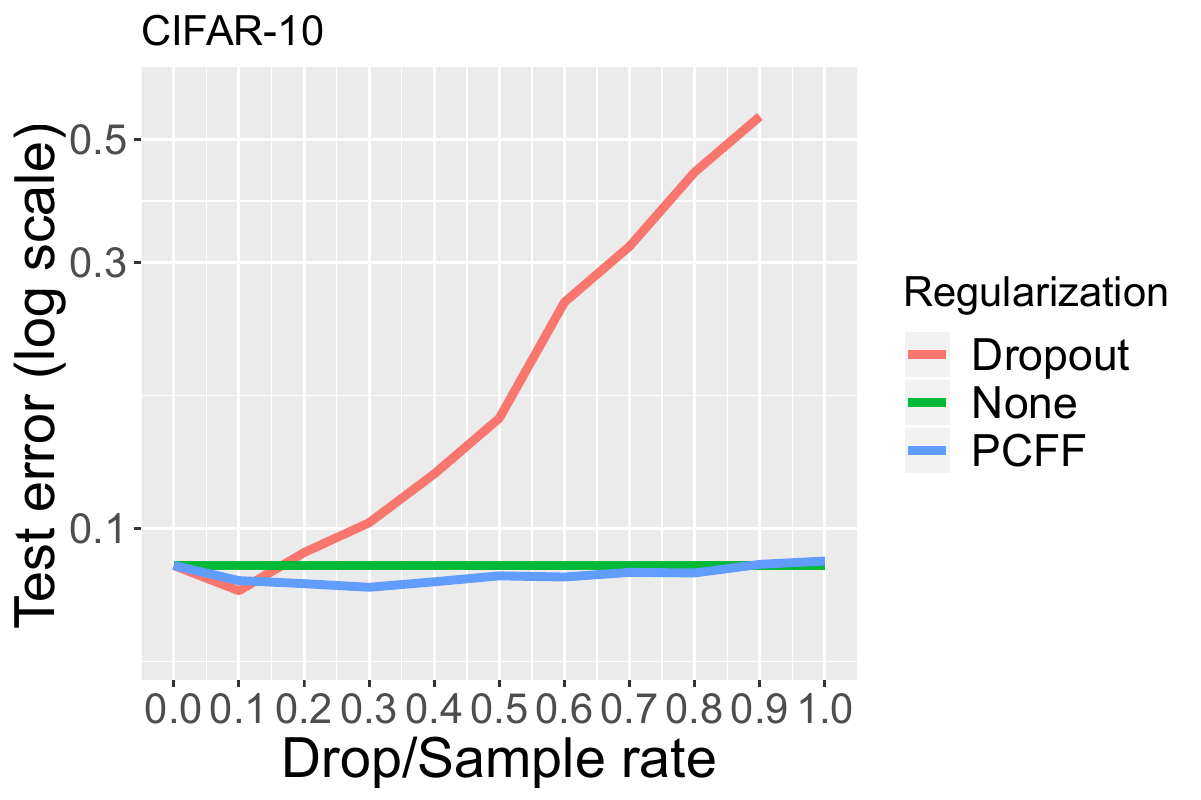}
\caption{Comparison of stochastic methods (None/Dropout/PCFF) in terms of image classification test errors (lower is better) under various settings. Left: MNIST/FashionMNIST datasets with a simple dense network and tanh/ReLU activation functions; Right: CIFAR-10 dataset with ResNet20 and varying drop/sample rates. All reported results are average values of three runs. Compared to dropout, PCFF can achieve comparable results, and tend to deliver more consistent improvements. } \label{fig:pcffexp}
\end{figure}

\paragraph{Simple dense network}
We start with a simple network with two dense hidden layers of 1024 nodes to classify MNIST \citep{lucun1998mnist} and FashionMNIST \citep{xiao2017fashionmnist} images. We use PyTorch \citep{paszke2017pytorch}, train with stochastic gradient descent (learning rate $0.01$, momentum $0.9$), and set up 20\% of training data as validation set for performance monitoring and early-stopping. We set drop rate to 0.5 for dropout, and for PCFF we set the sample rate to 0.4 for tanh and 1.0 (full sampling) for ReLU. Figure~\ref{fig:pcffexp}~Left reports the test errors with different activation functions and stochastic regularizations.

We see that dropout and PCFF are overall comparable, and both improve the results in most cases. Also, the ReLU activation consistently produces better results that tanh. Additional experiments show that PCFF and dropout can be used together, which sometimes yields improved performance.

\paragraph{Convolutional residual network}
To figure out the applicability of PCFF in convolutional residual networks, we experiment on CIFAR-10 \citep{krizhevsky09cifar} image classification. For this we adapt an existing implementation 
\citep{Idelbayev18a} to use the preactivation variant. We focus on the ResNet20 structure, and follow the original learning rate schedule except for setting up a validation set of 10\% training data to monitor training performance. Figure~\ref{fig:pcffexp}~Right summarizes the test errors under different drop/sample rates.

We observe that in this case PCFF can improve the performance over a wide range of sample rates, whereas dropout is only effective with drop rate $0.1$, and large drop rates in this case significantly deteriorate the performance. We also observe a clear trade-off of the PCFF sample rate, where a partial sampling of 0.3 yields the best result. 

\paragraph{Independently RNN}
We complete our empirical evaluations of PCFF with an RNN test case. For this we used IndRNNs with 6 layers to solve the sequential/permuted MNIST classification problems based on an existing Implementation\footnote{\url{https://github.com/Sunnydreamrain/IndRNN_pytorch}} provided by the authors of IndRNN \citep{li2018indrnn,li2019indrnn}. We tested over dropout with drop rate 0.1 and PCFF with sample rate 0.1 and report the average test accuracy of three runs. We notice that, while in the permuted MNIST case both dropout (0.9203) and PCFF (0.9145) improves the result (0.9045), in the sequential MNIST case, dropout (0.9830) seems to worsen the performance (0.9841) whereas PCFF (0.9842) delivers comparable result.

\section{Conclusions and discussions}

In this work, we show that neural networks can be interpreted as layered chain graphs, and that feed-forward can be viewed as an approximate inference procedure for these models. This chain graph interpretation provides a unified theoretical framework that elucidates the underlying mechanism of real-world neural networks and provides coherent and in-depth theoretical support for a wide range of empirically established network designs. Furthermore, it also offers a solid foundation to derive new deep learning approaches, with the additional help from the rich existing work on PGMs. It is thus a promising alternative neural network interpretation that deepens our theoretical understanding and unveils a new perspective for future deep learning research. 

In the future, we plan to investigate a number of open questions that stem from this work, especially:

\begin{itemize}
\item Is the current chain graph interpretation sufficient to capture the full essence of neural networks? Based on the current results, we are reasonably optimistic that the proposed interpretation can cover an essential part of the neural network mechanism. However, compared to the function approximator view, it only covers a subset of existing techniques. Is this subset good enough?
\item On a related note: can we find chain graph interpretations for other important network designs (or otherwise some chain graph interpretable alternatives with comparable or better performance)? The current work provides a good start, but it is by no means an exhaustive study.
\item Finally, what other new deep learning models and procedures can we build up based on the chain graph framework? The partially collapsed feed-forward inference proposed in this work is just a simple illustrative example, and we believe that many other promising deep learning techniques can be derived from the proposed chain graph interpretation.
\end{itemize}

\bibliography{iclr2021_conference}

\begin{thebibliography}{48}
\providecommand{\natexlab}[1]{#1}
\providecommand{\url}[1]{\texttt{#1}}
\expandafter\ifx\csname urlstyle\endcsname\relax
  \providecommand{\doi}[1]{doi: #1}\else
  \providecommand{\doi}{doi: \begingroup \urlstyle{rm}\Url}\fi

\bibitem[Bach et~al.(2015)Bach, Binder, Montavon, Klauschen, M{\"u}ller, and
  Samek]{bach2015lrp}
Sebastian Bach, Alexander Binder, Gr{\'e}goire Montavon, Frederick Klauschen,
  Klaus-Robert M{\"u}ller, and Wojciech Samek.
\newblock On pixel-wise explanations for non-linear classifier decisions by
  layer-wise relevance propagation.
\newblock \emph{PLOS ONE}, 10\penalty0 (7):\penalty0 1--46, 07 2015.

\bibitem[Buntine(1994)]{buntine1994gmasnn}
Wray~L. Buntine.
\newblock Operations for learning with graphical models.
\newblock \emph{J. Artif. Int. Res.}, 2\penalty0 (1):\penalty0 159--225,
  December 1994.

\bibitem[Chen et~al.(2018)Chen, Papandreou, Kokkinos, Murphy, and
  Yuille]{chen2018deeplab}
Liang{-}Chieh Chen, George Papandreou, Iasonas Kokkinos, Kevin Murphy, and
  Alan~L. Yuille.
\newblock Deeplab: Semantic image segmentation with deep convolutional nets,
  atrous convolution, and fully connected crfs.
\newblock \emph{TPAMI}, 40\penalty0 (4):\penalty0 834--848, 2018.

\bibitem[Chizat \& Bach(2018)Chizat and Bach]{chizat2018global}
Lenaic Chizat and Francis Bach.
\newblock On the global convergence of gradient descent for over-parameterized
  models using optimal transport.
\newblock In \emph{NeurIPS}, 2018.

\bibitem[Cho et~al.(2014)Cho, van Merri{\"e}nboer, Gulcehre, Bahdanau,
  Bougares, Schwenk, and Bengio]{cho2014gru}
Kyunghyun Cho, Bart van Merri{\"e}nboer, Caglar Gulcehre, Dzmitry Bahdanau,
  Fethi Bougares, Holger Schwenk, and Yoshua Bengio.
\newblock Learning phrase representations using {RNN} encoder{--}decoder for
  statistical machine translation.
\newblock In \emph{EMNLP}, 2014.

\bibitem[Devlin et~al.(2019)Devlin, Chang, Lee, and Toutanova]{devlin2019bert}
Jacob Devlin, Ming-Wei Chang, Kenton Lee, and Kristina Toutanova.
\newblock {BERT}: Pre-training of deep bidirectional transformers for language
  understanding.
\newblock In \emph{NAACL}, 2019.

\bibitem[Erhan et~al.(2009)Erhan, Bengio, Courville, and
  Vincent]{erhan2009visualizing}
Dumitru Erhan, Yoshua Bengio, Aaron Courville, and Pascal Vincent.
\newblock Visualizing higher-layer features of a deep network.
\newblock 2009.

\bibitem[Genevay et~al.(2017)Genevay, Peyr{\'e}, and Cuturi]{genevay2017gan}
Aude Genevay, Gabriel Peyr{\'e}, and Marco Cuturi.
\newblock {GAN} and {VAE} from an optimal transport point of view.
\newblock \emph{arXiv preprint arXiv:1706.01807}, 2017.

\bibitem[Goodfellow et~al.(2014)Goodfellow, Pouget-Abadie, Mirza, Xu,
  Warde-Farley, Ozair, Courville, and Bengio]{goodfellow2014gan}
I.~J. Goodfellow, J.~Pouget-Abadie, M.~Mirza, B.~Xu, D.~Warde-Farley, S.~Ozair,
  A.~Courville, and Y.~Bengio.
\newblock Generative adversarial networks.
\newblock In \emph{NIPS}, 2014.

\bibitem[Goodfellow et~al.(2015)Goodfellow, Shlens, and
  Szegedy]{goodfellow2015adversarial}
Ian Goodfellow, Jonathon Shlens, and Christian Szegedy.
\newblock Explaining and harnessing adversarial examples.
\newblock In \emph{ICLR}, 2015.

\bibitem[Goodfellow et~al.(2016)Goodfellow, Bengio, and
  Courville]{goodfellow2016dlbook}
Ian Goodfellow, Yoshua Bengio, and Aaron Courville.
\newblock \emph{Deep Learning}.
\newblock MIT Press, 2016.

\bibitem[{He} et~al.(2016){He}, {Zhang}, {Ren}, and {Sun}]{he2016resnet}
K.~{He}, X.~{Zhang}, S.~{Ren}, and J.~{Sun}.
\newblock Deep residual learning for image recognition.
\newblock In \emph{CVPR}, 2016.

\bibitem[He et~al.(2016)He, Zhang, Ren, and Sun]{he2016resnet2}
Kaiming He, Xiangyu Zhang, Shaoqing Ren, and Jian Sun.
\newblock Identity mappings in deep residual networks.
\newblock In \emph{ECCV}, 2016.

\bibitem[{Hinton} et~al.(2012){Hinton}, {Deng}, {Yu}, {Dahl}, {Mohamed},
  {Jaitly}, {Senior}, {Vanhoucke}, {Nguyen}, {Sainath}, and
  {Kingsbury}]{hinton2012speech}
G.~{Hinton}, L.~{Deng}, D.~{Yu}, G.~E. {Dahl}, A.~{Mohamed}, N.~{Jaitly},
  A.~{Senior}, V.~{Vanhoucke}, P.~{Nguyen}, T.~N. {Sainath}, and
  B.~{Kingsbury}.
\newblock Deep neural networks for acoustic modeling in speech recognition: The
  shared views of four research groups.
\newblock \emph{IEEE Signal Processing Magazine}, 29\penalty0 (6):\penalty0
  82--97, 2012.

\bibitem[Hinton et~al.(2006)Hinton, Osindero, and Teh]{hinton2006dbn}
Geoffrey~E. Hinton, Simon Osindero, and Yee-Whye Teh.
\newblock A fast learning algorithm for deep belief nets.
\newblock \emph{Neural Comput.}, 18\penalty0 (7):\penalty0 1527--1554, 2006.

\bibitem[Hochreiter \& Schmidhuber(1997)Hochreiter and
  Schmidhuber]{hochreiter1997lstm}
Sepp Hochreiter and J\"{u}rgen Schmidhuber.
\newblock Long short-term memory.
\newblock \emph{Neural Comput.}, 9\penalty0 (8):\penalty0 1735--1780, November
  1997.

\bibitem[Idelbayev()]{Idelbayev18a}
Yerlan Idelbayev.
\newblock Proper {ResNet} implementation for {CIFAR10/CIFAR100} in {PyTorch}.
\newblock \url{https://github.com/akamaster/pytorch_resnet_cifar10}.

\bibitem[Ioffe \& Szegedy(2015)Ioffe and Szegedy]{ioffe2015batchnorm}
Sergey Ioffe and Christian Szegedy.
\newblock Batch normalization: Accelerating deep network training by reducing
  internal covariate shift.
\newblock In \emph{ICML}, 2015.

\bibitem[Jacot et~al.(2018)Jacot, Gabriel, and Hongler]{jacot2018ntk}
Arthur Jacot, Franck Gabriel, and Cl\'{e}ment Hongler.
\newblock Neural tangent kernel: Convergence and generalization in neural
  networks.
\newblock In \emph{NeurIPS}, 2018.

\bibitem[Jang et~al.(2017)Jang, Gu, and Poole]{jang2017gumbelsoftmax}
Eric Jang, Shixiang Gu, and Ben Poole.
\newblock Categorical reparameterization with gumbel-softmax.
\newblock In \emph{ICLR}, 2017.

\bibitem[Kingma \& Welling(2014)Kingma and Welling]{kingma2014vae}
Diederik~P. Kingma and Max Welling.
\newblock Auto-encoding variational bayes.
\newblock In \emph{ICLR}, 2014.

\bibitem[Koller \& Friedman(2009)Koller and Friedman]{koller2009pgmbook}
Daphne Koller and Nir Friedman.
\newblock \emph{Probabilistic Graphical Models: Principles and Techniques}.
\newblock MIT Press, 2009.

\bibitem[Krizhevsky et~al.(2012)Krizhevsky, Sutskever, and
  Hinton]{krizhevsky12alexnet}
A.~Krizhevsky, I.~Sutskever, and G.~E. Hinton.
\newblock Image{N}et classification with deep convolutional neural networks.
\newblock In \emph{NIPS}, 2012.

\bibitem[Krizhevsky(2009)]{krizhevsky09cifar}
Alex Krizhevsky.
\newblock Learning multiple layers of features from tiny images.
\newblock Technical report, 2009.

\bibitem[Lample et~al.(2016)Lample, Ballesteros, Subramanian, Kawakami, and
  Dyer]{lampe2016crfnlp}
Guillaume Lample, Miguel Ballesteros, Sandeep Subramanian, Kazuya Kawakami, and
  Chris Dyer.
\newblock Neural architectures for named entity recognition.
\newblock In \emph{NACCL}, 2016.

\bibitem[Lecun et~al.(1998)Lecun, Bottou, Bengio, and Haffner]{lucun1998mnist}
Y.~Lecun, L.~Bottou, Y.~Bengio, and P.~Haffner.
\newblock Gradient-based learning applied to document recognition.
\newblock \emph{Proceedings of the IEEE}, 86\penalty0 (11):\penalty0
  2278--2324, 1998.

\bibitem[Lee et~al.(2018)Lee, Bahri, Novak, Schoenholz, Pennington, and
  Sohl{-}Dickstein]{lee2018dnnasgp}
Jaehoon Lee, Yasaman Bahri, Roman Novak, Samuel~S. Schoenholz, Jeffrey
  Pennington, and Jascha Sohl{-}Dickstein.
\newblock Deep neural networks as gaussian processes.
\newblock In \emph{ICLR}, 2018.

\bibitem[Li et~al.(2018)Li, Li, Cook, Zhu, and Gao]{li2018indrnn}
Shuai Li, Wanqing Li, Chris Cook, Ce~Zhu, and Yanbo Gao.
\newblock Independently recurrent neural network (indrnn): Building a longer
  and deeper {RNN}.
\newblock In \emph{CVPR}, 2018.

\bibitem[Li et~al.(2019)Li, Li, Cook, Gao, and Zhu]{li2019indrnn}
Shuai Li, Wanqing Li, Chris Cook, Yanbo Gao, and Ce~Zhu.
\newblock Deep independently recurrent neural network (indrnn).
\newblock \emph{arXiv preprint arXiv:1910.06251}, 2019.

\bibitem[Lipton(2018)]{lipton2018mythos}
Zachary~C. Lipton.
\newblock The mythos of model interpretability.
\newblock \emph{Commun. ACM}, 61\penalty0 (10):\penalty0 36--43, September
  2018.

\bibitem[Long et~al.(2015)Long, Shelhamer, and Darrell]{long15fully}
Jonathan Long, Evan Shelhamer, and Trevor Darrell.
\newblock Fully convolutional networks for semantic segmentation.
\newblock In \emph{CVPR}, 2015.

\bibitem[Mei et~al.(2019)Mei, Misiakiewicz, and Montanari]{mei2019mean}
Song Mei, Theodor Misiakiewicz, and Andrea Montanari.
\newblock Mean-field theory of two-layers neural networks: dimension-free
  bounds and kernel limit.
\newblock \emph{Proceedings of Machine Learning Research}, \penalty0 (99),
  2019.

\bibitem[Mnih et~al.(2015)Mnih, Kavukcuoglu, Silver, Rusu, Veness, Bellemare,
  Graves, Riedmiller, Fidjeland, Ostrovski, Petersen, Beattie, Sadik,
  Antonoglou, King, Kumaran, Wierstra, Legg, and Hassabis]{mnih2015dqn}
Volodymyr Mnih, Koray Kavukcuoglu, David Silver, Andrei~A. Rusu, Joel Veness,
  Marc~G. Bellemare, Alex Graves, Martin Riedmiller, Andreas~K. Fidjeland,
  Georg Ostrovski, Stig Petersen, Charles Beattie, Amir Sadik, Ioannis
  Antonoglou, Helen King, Dharshan Kumaran, Daan Wierstra, Shane Legg, and
  Demis Hassabis.
\newblock Human-level control through deep reinforcement learning.
\newblock \emph{Nature}, 518\penalty0 (7540):\penalty0 529--533, February 2015.

\bibitem[Nair \& Hinton(2010)Nair and Hinton]{nair2010relu}
Vinod Nair and Geoffrey~E. Hinton.
\newblock Rectified linear units improve restricted boltzmann machines.
\newblock In \emph{ICML}, 2010.

\bibitem[Neal(1992)]{neal1992sbn}
Radford~M. Neal.
\newblock Connectionist learning of belief networks.
\newblock \emph{AI}, 56\penalty0 (10):\penalty0 71--113, July 1992.

\bibitem[Neal(1996)]{neal1996phd}
Radford~M. Neal.
\newblock \emph{Bayesian Learning for Neural Networks}.
\newblock Springer-Verlag, 1996.

\bibitem[Nguyen et~al.(2016)Nguyen, Dosovitskiy, Yosinski, Brox, and
  Clune]{nguyen2016synthinput}
Anh Nguyen, Alexey Dosovitskiy, Jason Yosinski, Thomas Brox, and Jeff Clune.
\newblock Synthesizing the preferred inputs for neurons in neural networks via
  deep generator networks.
\newblock In \emph{NIPS}, 2016.

\bibitem[Paszke et~al.(2017)Paszke, Gross, Chintala, Chanan, Yang, DeVito, Lin,
  Desmaison, Antiga, and Lerer]{paszke2017pytorch}
Adam Paszke, Sam Gross, Soumith Chintala, Gregory Chanan, Edward Yang, Zachary
  DeVito, Zeming Lin, Alban Desmaison, Luca Antiga, and Adam Lerer.
\newblock Automatic differentiation in pytorch.
\newblock In \emph{NIPS-W}, 2017.

\bibitem[Poon \& Domingos(2011)Poon and Domingos]{poon2011spn}
Hoifung Poon and Pedro Domingos.
\newblock Sum-product networks: A new deep architecture.
\newblock In \emph{UAI}, 2011.

\bibitem[Salakhutdinov \& Hinton(2012)Salakhutdinov and
  Hinton]{salakhutdinov2012dbm}
Ruslan Salakhutdinov and Geoffrey Hinton.
\newblock An efficient learning procedure for deep boltzmann machines.
\newblock \emph{Neural Comput.}, 24\penalty0 (8):\penalty0 1967--2006, August
  2012.

\bibitem[Shen et~al.(2019)Shen, Wu, Domokos, and Cremers]{shen2019lgm}
Yuesong Shen, Tao Wu, Csaba Domokos, and Daniel Cremers.
\newblock Probabilistic discriminative learning with layered graphical models.
\newblock \emph{CoRR}, abs/1902.00057, 2019.

\bibitem[Silver et~al.(2016)Silver, Huang, Maddison, Guez, Sifre, van~den
  Driessche, Schrittwieser, Antonoglou, Panneershelvam, Lanctot, Dieleman,
  Grewe, Nham, Kalchbrenner, Sutskever, Lillicrap, Leach, Kavukcuoglu, Graepel,
  and Hassabis]{silver2016alphago}
David Silver, Aja Huang, Chris~J. Maddison, Arthur Guez, Laurent Sifre, George
  van~den Driessche, Julian Schrittwieser, Ioannis Antonoglou, Veda
  Panneershelvam, Marc Lanctot, Sander Dieleman, Dominik Grewe, John Nham, Nal
  Kalchbrenner, Ilya Sutskever, Timothy Lillicrap, Madeleine Leach, Koray
  Kavukcuoglu, Thore Graepel, and Demis Hassabis.
\newblock Mastering the game of {Go} with deep neural networks and tree search.
\newblock \emph{Nature}, 529\penalty0 (7587):\penalty0 484--489, January 2016.

\bibitem[Srivastava et~al.(2014)Srivastava, Hinton, Krizhevsky, Sutskever, and
  Salakhutdinov]{srivastava2014dropout}
Nitish Srivastava, Geoffrey Hinton, Alex Krizhevsky, Ilya Sutskever, and Ruslan
  Salakhutdinov.
\newblock Dropout: A simple way to prevent neural networks from overfitting.
\newblock \emph{JMLR}, 15\penalty0 (56):\penalty0 1929--1958, 2014.

\bibitem[{Szegedy} et~al.(2015){Szegedy}, {Wei Liu}, {Yangqing Jia},
  {Sermanet}, {Reed}, {Anguelov}, {Erhan}, {Vanhoucke}, and
  {Rabinovich}]{szegedy2015googlenet}
C.~{Szegedy}, {Wei Liu}, {Yangqing Jia}, P.~{Sermanet}, S.~{Reed},
  D.~{Anguelov}, D.~{Erhan}, V.~{Vanhoucke}, and A.~{Rabinovich}.
\newblock Going deeper with convolutions.
\newblock In \emph{CVPR}, 2015.

\bibitem[Vaswani et~al.(2017)Vaswani, Shazeer, Parmar, Uszkoreit, Jones, Gomez,
  Kaiser, and Polosukhin]{vaswani2017attention}
Ashish Vaswani, Noam Shazeer, Niki Parmar, Jakob Uszkoreit, Llion Jones,
  Aidan~N Gomez, {\L}ukasz Kaiser, and Illia Polosukhin.
\newblock Attention is all you need.
\newblock In \emph{NIPS}, 2017.

\bibitem[Xiao et~al.(2017)Xiao, Rasul, and Vollgraf]{xiao2017fashionmnist}
Han Xiao, Kashif Rasul, and Roland Vollgraf.
\newblock Fashion-mnist: a novel image dataset for benchmarking machine
  learning algorithms.
\newblock \emph{CoRR}, abs/1708.07747, 2017.

\bibitem[Zeiler \& Fergus(2014)Zeiler and Fergus]{zeiler2014deconv}
Matthew~D. Zeiler and Rob Fergus.
\newblock Visualizing and understanding convolutional networks.
\newblock In \emph{ECCV}, 2014.

\bibitem[Zheng et~al.(2015)Zheng, Jayasumana, Romera{-}Paredes, Vineet, Su, Du,
  Huang, and Torr]{zheng2015crfasrnn}
Shuai Zheng, Sadeep Jayasumana, Bernardino Romera{-}Paredes, Vibhav Vineet,
  Zhizhong Su, Dalong Du, Chang Huang, and Philip H.~S. Torr.
\newblock Conditional random fields as recurrent neural networks.
\newblock In \emph{ICCV}, 2015.

\end{thebibliography}
\bibliographystyle{iclr2021_conference}

\appendix

\section{Proofs}

\subsection{Proof of Proposition~\ref{thm:ff}} \label{asubsec:proof_thm_ff}

\begin{proof}[\unskip\nopunct]
The main idea behind the proof is that for a linear function, its expectation can be moved inside and directly applied on its arguments. With this in mind let's start the actual deductions:
\begin{itemize}
\item To obtain Eq.~\eqref{eq:distff}, we start from Eqs.~\eqref{eq:qrecursive} and \eqref{eq:activgeneral}:
\begin{align}
Q^l_i (x^l_i | \tilde{\bfx}^1) &= \bbE_{Q^{p_1},\dots,Q^{p_n}}[P(x^l_i | \pa(\bfX^l))] \\
&= \bbE_{Q^{p_1},\dots,Q^{p_n}}[f^{l} \big( \bfT^l(X^{l}_i), \bfe^{l}_i\big(\bfT^{p_1}(\bfX^{p_1}),\dots,\bfT^{p_n}(\bfX^{p_n})\big) \big)].
\end{align}
Now, we make the assumption that the following mapping is approximately linear:
\begin{equation}
(\bfv_1, \dots,\bfv_n) \mapsto f^{l} \big( \bfT^l(X^{l}_i), \bfe^{l}_i(\bfv_1, \dots,\bfv_n) \big).
\end{equation}
This allows us to move the expectation inside, resulting in (racall the definition of $\bfq$ in Eq.~\eqref{eq:qdefs})
\begin{align}
Q^l_i (x^l_i | \tilde{\bfx}^1) &\approx f^{l} \big( \bfT^l(X^{l}_i), \bfe^{l}_i\big(\bbE_{Q^{p_1}}[\bfT^{p_1}(\bfX^{p_1})],\dots,\bbE_{Q^{p_n}}[\bfT^{p_n}(\bfX^{p_n})]\big) \big) \\
&\approx f^{l} \big( \bfT^l(X^{l}_i), \bfe^{l}_i(\bfq^{p_1},\dots,\bfq^{p_n}) \big).
\end{align}
\item To obtain Eq.~\eqref{eq:ff}, we go through a similar procedure (for discrete RVs we replace integrations by summations):

From Eqs.~\eqref{eq:qdefs}, \eqref{eq:qrecursive} and \eqref{eq:activgeneral} we have
\begin{align}
\bfq^l_i &= \bbE_{Q^l_i}[\bfT^l(X^l_i)] \\
&= \int_{x^l_i} \bfT^l(x^l_i) Q^l_i (x^l_i | \tilde{\bfx}^1) dx^l_i \\
&= \int_{x^l_i} \bfT^l(x^l_i) \bbE_{Q^{p_1},\dots,Q^{p_n}}[f^{l} \big( \bfT^l(x^{l}_i), \bfe^{l}_i\big(\bfT^{p_1}(\bfX^{p_1}),\dots,\bfT^{p_n}(\bfX^{p_n})\big) \big)] dx^l_i \\
&= \bbE_{Q^{p_1},\dots,Q^{p_n}}\Big[\int_{x^l_i} \bfT^l(x^l_i) f^{l} \big( \bfT^l(x^{l}_i), \bfe^{l}_i\big(\bfT^{p_1}(\bfX^{p_1}),\dots,\bfT^{p_n}(\bfX^{p_n})\big) \big) dx^l_i\Big].
\end{align}
Defining the activation function $\bfg^l$ as (thus $\bfe^{l}_i$ corresponds to the preactivation)
\begin{equation}
\bfg^l (\bfv) = \int_{x^l_i} \bfT^l(x^l_i) f^{l} \big( \bfT^l(x^{l}_i), \bfv \big) dx^l_i,
\end{equation}
we then have
\begin{equation}
\bfq^l_i = \bbE_{Q^{p_1},\dots,Q^{p_n}}\big[\bfg^{l} \big(\bfe^{l}_i\big(\bfT^{p_1}(\bfX^{p_1}),\dots,\bfT^{p_n}(\bfX^{p_n})\big) \big)\big].
\end{equation}

Again, we make another assumption that the following mapping is approximately linear:
\begin{equation}
(\bfv_1, \dots,\bfv_n) \mapsto \bfg^{l} \big( \bfe^{l}_i(\bfv_1, \dots,\bfv_n) \big).
\end{equation}
This leads to the following approximation in a similar fashion:
\begin{align}
\bfq^l_i &\approx \bfg^{l} \big( \bfe^{l}_i\big(\bbE_{Q^{p_1}}[\bfT^{p_1}(\bfX^{p_1})],\dots,\bbE_{Q^{p_n}}[\bfT^{p_n}(\bfX^{p_n})]\big) \big) \\
&\approx \bfg^{l} \big( \bfe^{l}_i(\bfq^{p_1},\dots,\bfq^{p_n}) \big).
\end{align}
\end{itemize}
\end{proof}

\subsection{Proof (and other details) of Corollary~\ref{cor:activationfuncs}} \label{asubsec:proof_cor_activationfuncs}

\begin{proof}[\unskip\nopunct]

Let's consider a node $X^l_i$ connected by parent layers $\bfX^{p_1},\dots,\bfX^{p_n}$. To lessen the notations we use the shorthands $\bfe^{l}_i$ for $\bfe^{l}_i\big(\bfT^{p_1}(\bfX^{p_1}),\dots,\bfT^{p_n}(\bfX^{p_n})\big)$ and $\bar{\bfe}^{l}_i$ for $\bfe^{l}_i(\bfq^{p_1},\dots,\bfq^{p_n})$.

\begin{enumerate}
\item For the binary case we have
\begin{align}
T^l(X^l_i) &= X^l_i \in \{ \alpha, \beta \}, \\
P(X^{l}_i|\pa(\bfX^l)) &= f^{l} ( X^{l}_i, e^{l}_i ) = \frac{1}{Z(\pa(\bfX^l))} \exp(X^{l}_i \ e^{l}_i)
\end{align}
with the partition function
\begin{equation}
Z(\pa(\bfX^l)) = \exp( \alpha \ e^{l}_i ) + \exp( \beta \ e^{l}_i)
\end{equation}
that makes sure $P(X^{l}_i|\pa(\bfX^l))$ is normalized. This means that, since $X^l_i$ can either be $\alpha$ or $\beta$, we can equivalently write ($\sigma: x \mapsto 1 / (1 + \exp(-x))$ denotes the sigmoid function)
\begin{equation}
f^{l} ( X^{l}_i, e^{l}_i ) = P(x^{l}_i|\pa(\bfx^l)) = 
\begin{cases}
\sigma ( (\alpha-\beta) \ e^{l}_i) & \text{if } x^{l}_i = \alpha \\
\sigma ( (\beta-\alpha) \ e^{l}_i) & \text{if } x^{l}_i = \beta
\end{cases}
= \sigma( (2 x^{l}_i - \alpha - \beta) \ e^{l}_i).
\end{equation}
Using the feed-forward expression (Eq.~\eqref{eq:ff}), we have
\begin{align}
q^l_i \approx \sum_{x^l_i \in \{ \alpha, \beta \}} x^l_i \ f^{l} ( x^{l}_i, \bar{e}^{l}_i ) &= \alpha \ \sigma ( (\alpha-\beta) \ \bar{e}^{l}_i) + \beta \ \sigma ( (\beta-\alpha) \ \bar{e}^{l}_i) \\
&= \frac{\beta-\alpha}{2} \tanh \Big(\frac{\beta-\alpha}{2} \cdot \bar{e}^{l}_i \Big) + \frac{\alpha+\beta}{2}.
\end{align}
Especially,
\begin{align}
\text{When }\alpha=0, \beta=1,&\text{ we have } q^l_i \approx \sigma (\bar{e}^{l}_i); \\
\text{When }\alpha=-1, \beta=1,&\text{ we have } q^l_i \approx \tanh (\bar{e}^{l}_i).
\end{align}
Furthermore, the roles of $\alpha$ and $\beta$ are interchangeable.

\item For the multilabel case, let's assume that the node $X^l_i$ can take one of the $c$ labels $\{1, \dots, c\}$. In this case, we have an indicator feature function which outputs a length-$c$ feature vector
\begin{equation}
\bfT^l(X^l_i) = (\indifunc{X^l_i=1}, \dots, \indifunc{X^l_i=c})^{\top}.
\end{equation} 
This means that for any given label $j \in \{1, \dots, c\}$, $\bfT^l(j)$ is a one-hot vector indicating the $j$-th position. Also, $\bfe^l_i$ and $\bar{\bfe}^{l}_i$ will both be vectors of length $c$, and we denote $e^l_{i, j}$ and $\bar{e}^{l}_{i,j}$ their $j$-th entries. We have then
\begin{equation}
P(X^{l}_i|\pa(\bfX^l)) = f^{l} ( \bfT^l(X^{l}_i), \bfe^{l}_i ) = \frac{1}{Z(\pa(\bfX^l))} \exp(\bfT^l(X^{l}_i) \cdot \bfe^{l}_i)
\end{equation}
with the normalizer (i.e.\ partition function)
\begin{equation}
Z(\pa(\bfX^l)) = \sum_{j=1}^c \exp( e^{l}_{i,j} ).
\end{equation}
This means that
\begin{equation}
\forall j \in \{1, \dots, c\}, f^{l} ( \bfT^l(j), \bfe^{l}_i ) = \big( \softmax(e^{l}_{i,1}, \dots, e^{l}_{i,c}) \big)_j,
\end{equation}
and, using the feed-forward expression (Eq.~\eqref{eq:ff}), we have
\begin{align}
\bfq^l_i \approx \sum_{x^l_i = 1}^c f^{l} ( \bfT^l(x^l_i), \bar{\bfe}^{l}_i ) \bfT^l(x^l_i) &= \sum_{j = 1}^c \big( \softmax(\bar{e}^{l}_{i,1}, \dots, \bar{e}^{l}_{i,c}) \big)_j \bfT^l(j) \\
&= \softmax(\bar{e}^{l}_{i,1}, \dots, \bar{e}^{l}_{i,c}),
\end{align}
i.e.\ the expected features $\bfq^l_i$ of the multi-labeled node $X^l_i$ is a length-$c$ vector that encodes the result of a $\softmax$ activation.

\item The analytical forms of the activation functions are quite complicated for rectified Gaussian nodes. Luckily, it is straight-forward to sample from rectified Gaussian distributions (get Gaussian samples, then rectify). meaning that we can easily evaluate them numerically with sample averages. A resulting visualization is displayed in Figure~\ref{fig:banner}~Right. Specifically:

\begin{itemize}
	\item The ReLU nonlinearity can be approximated reasonably well by a rectified Gaussian node with no leak ($\epsilon=0$) and $\tanh$-modulated standard deviation ($s^l_i = \tanh(e^l_i)$), as shown by the red plot in Figure~\ref{fig:banner}~Right;
	\item Similar to the ReLU case, the leaky ReLU nonlinearity can be approximated by a leaky ($\epsilon \neq 0$) rectified Gaussian node with $\tanh$-modulated standard deviation ($s^l_i = \tanh(e^l_i)$). See the green plot in Figure~\ref{fig:banner}~Right which depict the case with leaky factor $\epsilon = 1 / 3$;
	\item We discover that a rectified Gaussian node with no leak ($\epsilon=0$) and an appropriately-chosen constant standard deviation $s^l_i$ can closely approximate the softplus nonlinearity (see the blue plot in Figure~\ref{fig:banner}~Right). We numerically evaluate $s^l_i=1.776091849725427$ to minimize the maximum pointwise approximation error.
\end{itemize}

Averaging over more samples would of course lead to more accurate (visually thinner) plots, however  in Figure~\ref{fig:banner}~Right we deliberately only average over 200 samples, because we also want to visualize their stochastic behaviors: the perceived thickness of a plot can provide a hint to the output sample variance given the preactivation $e^l_i$.
\end{enumerate}
\end{proof}

\subsection{Proof of Proposition \ref{prop:residual}} \label{asubsec:proof_prop_residual}

\begin{proof}[\unskip\nopunct]
Given a refinement module (c.f.\ Definition~\ref{def:refinementmodule}) that augments a base submodule $m$ from layer $\bfX^{l-1}$ to layer $\bfX^l$ using a refining submodule $r$ from $\tilde{\bfX}^{l}$ to layer $\bfX^l$, denote $\bfg^l$ the activation function corresponding to the distribution of nodes in layer $\bfX^l$, we assume that these two submodules alone would represent the following mappings during feed-forward
\begin{equation}
\begin{cases}
\bfq^l = \bfg^l(\bfe^m(\bfq^{l-1})) & \text{ (base submodule)}\\
\bfq^l = \bfg^l(\bfe^r(\tilde{\bfq}^{l})) & \text{ (refining submodule)}
\end{cases}
\end{equation}
where $\bfe^m$ and $\bfe^r$ represent the output preactivations of the base submodule and the refining submodule respectively. Then, given an input activation $\bfq^{l-1}$ from $\bfX^{l-1}$, the output preactivation of the overall refinement module should sum up contributions from both the main and the side branches (c.f.\ Eq.~\eqref{eq:activgeneralpreact}), meaning that the refinement module computes the output as
\begin{equation}
\bfq^l = \bfg^l( \bfe^m(\bfq^{l-1}) + \bfe^r(\tilde{\bfq}^{l}) )
\end{equation}
with $\tilde{\bfq}^{l}$ the output of the duplicated base submodule with shared weight, given by
\begin{equation}
\tilde{\bfq}^{l} = \bfg^l(\bfe^m(\bfq^{l-1})).
\end{equation}
We have thus ($\id$ denotes the identity function)
\begin{equation}
\bfq^l = \bfg^l \circ (\id + \bfe^r \circ \bfg^l) \circ \bfe^m(\bfq^{l-1})
\end{equation}
where the function $\id + \bfe^r \circ \bfg^l$ describes a preactivation residual block that arises naturally from the refinement module structure.
\end{proof}

\subsection{Proof of Proposition \ref{prop:rnn}} \label{asubsec:proof_prop_rnn}

\begin{proof}[\unskip\nopunct]
To match the typical deep learning formulations and ease the derivation, we assume that the base layered chain graph has a sequential structure, meaning that $\pa^t(\bfX^{l,t})$ contains only $\bfX^{l-1,t}$, and we have
\begin{align}
P(X^{l,t}_i|\pa^t(\bfX^{l,t}), \bfX^{l,t-1}) &= P(X^{l,t}_i|\bfX^{l-1,t}, \bfX^{l,t-1}) \\
&= f^{l,t}\big( \bfT^l(X^{l,t}_i), \bfe^{l,t}_i \big( \bfT^{l-1}(\bfX^{l-1,t}), \bfT^{l}(\bfX^{l,t-1}) \big) \big).
\end{align}
\begin{enumerate}
\item When the connection from $\bfX^{l,t-1}$ to $\bfX^{l,t}$ is dense, we have that for each $i \in \{1,\dots,N^{l}\}$, 
\begin{equation}
\bfe^{l,t}_i \big( \bfT^{l-1}(\bfX^{l-1,t}), \bfT^{l}(\bfX^{l,t-1}) \big) = \sum_{j=1}^{N^{l-1}} \bfW^l_{j, i} \bfT^{l-1}(X^{l-1,t}_j)+ \sum_{k=1}^{N^{l}} \bfU^l_{k, i} \bfT^{l}(X^{l,t-1}_k) + \mathbf{b}^{l}_i.
\end{equation}
Thus the feed-forward update for layer $l$ at time $t$ is
\begin{equation}
\forall i \in \{1,\dots,N^{l}\}, \ \bfq^{l,t}_i \approx \bfg^l \big( \sum_{j=1}^{N^{l-1}} \bfW^l_{j, i} \bfq^{l-1,t}_j + \sum_{k=1}^{N^{l}} \bfU^l_{k, i} \bfq^{l,t-1}_k + \mathbf{b}^{l}_i \big)
\end{equation}
which corresponds to the update of a simple recurrent layer.
\item The assumption of intra-layer conditional independence through time means that we have
\begin{equation}
P(X^{l,t}_i|\pa^t(\bfX^{l,t}), \bfX^{l,t-1}) = P(X^{l,t}_i|\bfX^{l-1,t}, X^{l,t-1}_i),
\end{equation}
which in terms of preactivation function means that for each $i \in \{1,\dots,N^{l}\}$, 
\begin{align}
\bfe^{l,t}_i \big( \bfT^{l-1}(\bfX^{l-1,t}), \bfT^{l}(\bfX^{l,t-1}) \big) &= \bfe^{l,t}_i \big( \bfT^{l-1}(\bfX^{l-1,t}), \bfT^{l}(X^{l,t-1}_i) \big)\\
&= \sum_{j=1}^{N^{l-1}} \bfW^l_{j, i} \bfT^{l-1}(X^{l-1,t}_j)+ \bfU^l_{i} \bfT^{l}(X^{l,t-1}_i) + \mathbf{b}^{l}_i.
\end{align}
In this case the feed-forward update for layer $l$ at time $t$ is
\begin{equation}
\forall i \in \{1,\dots,N^{l}\}, \ \bfq^{l,t}_i \approx \bfg^l \big( \sum_{j=1}^{N^{l-1}} \bfW^l_{j, i} \bfq^{l-1,t}_j + \bfU^l_{i} \bfq^{l,t-1}_i + \mathbf{b}^{l}_i \big)
\end{equation}
which corresponds to the update of an independently RNN layer (c.f.\ Eq.~(2) of \citet{li2018indrnn}).
\end{enumerate}
\end{proof}

\subsection{Proof of Proposition \ref{prop:dropout}} \label{asubsec:proof_prop_dropout}

\begin{proof}[\unskip\nopunct]
Again, to match the typical deep learning formulations and ease the derivation, we assume that the layered chain graph has a sequential structure, meaning that $\bfX^{l}$ is only the parent layer of $\bfX^{l+1}$. With the introduction of the auxiliary Bernoulli RVs $\bfD^l$, the $l+1$-th chain component represents
\begin{equation}
P(\bfX^{l+1}|\bfX^{l}, \bfD^{l}) = \prod_{j=1}^{N^{l+1}} f^{l+1}_j \big( \bfT^{l+1}(X^{l+1}_j), \bfe^{l+1}_j \big( \bfT^{l}(\bfX^{l}), \bfD^{l} \big) \big)
\end{equation}
with
\begin{equation}
\bfe^{l+1}_j \big( \bfT^{l}(\bfX^{l}), \bfD^{l} \big) = \sum_{i=1}^{N^{l-1}} D^l_i \bfW^{l+1}_{i, j} \bfT^{l}(X^{l}_i) + \mathbf{b}^{l}_j.
\end{equation}
\begin{itemize}
\item For a feed-forward pass during training, we draw a sample $d^l_i$ for each Bernoulli RV $D^l_i$, and the feed-forward update for layer $l+1$ becomes
\begin{equation}
\forall j \in \{1,\dots,N^{l+1}\}, \ \bfq^{l+1}_j \approx \bfg^{l+1} \big( \sum_{i=1}^{N^{l}} d^l_i \bfW^{l+1}_{i, j} \bfq^{l}_i + \mathbf{b}^{l}_j \big).
\end{equation}
Since $d^l_i=1$ with probability $p^l$ and $d^l_i=0$ with probability $1-p^l$, each activation $\bfq^{l}_i$ will be ``dropped out'' (i.e.\ $ d^l_i \bfq^{l}_i = 0$) with probability $1-p^l$ (this affects all $\bfq^{l+1}_j$ simultaneously). 
\item For a feed-forward pass at test time, we marginalize each  Bernoulli RV $D^l_i$. Since we have
\begin{equation}
\forall i \in \{1,\dots,N^l\}, \ \bbE[D^l_i] = p^l,
\end{equation}
the feed-forward update for layer $l+1$ in this case becomes
\begin{equation}
\forall j \in \{1,\dots,N^{l+1}\}, \ \bfq^{l+1}_j \approx \bfg^{l+1} \big( \sum_{i=1}^{N^{l}} p^l \bfW^{l+1}_{i, j} \bfq^{l}_i + \mathbf{b}^{l}_j \big)
\end{equation}
where we see the appearance of the constant scale $p^l$.
\end{itemize}
\end{proof}

\section{Remark on input modeling}

A technical detail that was not elaborated in the discussion of chain graph interpretation (Section~\ref{sec:nnascg}) is the input modeling: How to encode an input data sample? Ordinarily, an input data sample is treated as a sample drawn from some distribution that represents the input. In our case however, since the feed-forward process only pass through feature expectations, we can also directly interpret an input data sample as a feature expectation, meaning as a resulting average rather than a single sample. Using this fact, \citet{shen2019lgm} propose the ``soft clamping'' approach to encode real valued input taken from an interval, such as pixel intensity, simply as an expected value of a binary node which chooses between the interval boundary values. 

This said, since only the conditional distribution $P(\bfX^2, \dots, \bfX^L | \bfX^1)$  is modeled, our discriminative setting actually do not require specifying an input distribution $P(\bfX^1)$.

\end{document}